\documentclass{article} % For LaTeX2e
\usepackage{iclr2027_conference,times}

\usepackage{amsmath,amsfonts,bm}

\def\eqref#1{equation~\ref{#1}}
\def\1{\bm{1}}

\DeclareMathAlphabet{\mathsfit}{\encodingdefault}{\sfdefault}{m}{sl}
\SetMathAlphabet{\mathsfit}{bold}{\encodingdefault}{\sfdefault}{bx}{n}

\usepackage{hyperref}
\usepackage{url}

\usepackage{graphicx} 
\usepackage{booktabs}
\usepackage{capt-of}
\usepackage{algorithm}
\usepackage{multirow}
\usepackage{amsmath}
\usepackage{algpseudocode}
\usepackage{amssymb}

\title{RankBuffer: Efficient Ranking-Based Rewards for Open-Ended Generation}

\author{
\textbf{Zixuan Yang}$^{1}$\thanks{Equal contribution.} \quad
\textbf{Yiqun Chen}$^{1}$\footnotemark[1] \quad
\textbf{Qi Liu}$^{1}$ \quad
\textbf{Wei Yang}$^{2}$ \quad
\textbf{Erhan Zhang}$^{3}$ \\
\textbf{Liyi Chen}$^{3}$ \quad
\textbf{Qimeng Wang}$^{3}$ \quad
\textbf{Yan Gao}$^{3}$ \quad
\textbf{Jiaxin Mao}$^{1}$\thanks{Corresponding author.} \\
$^{1}$Renmin University of China, \quad
$^{2}$University of Southern California, \quad
$^{3}$Xiaohongshu Inc. \\
\texttt{zixuanyang@ruc.edu.cn} \quad
\texttt{maojiaxin@gmail.com}
}

\iclrfinalcopy % Uncomment for camera-ready version, but NOT for submission.
\begin{document}

\maketitle
\lhead{Under review as a conference paper at ICLR 2027}

\begin{abstract}
Open-ended generation lacks canonical answers, making pointwise rewards
difficult to calibrate for group-based reinforcement learning. Directly
ranking same-query rollouts provides a more suitable relative reward signal,
but existing ranking-based reward methods can incur substantial judging cost.
We introduce \textbf{RankBuffer}, which maintains an ordered, query-specific
buffer of previously judged responses as a reusable quality scale. Each
rollout is first inserted into an anchor interval through an independent
coarse judgment, after which only rollouts assigned to the same interval
undergo local fine ranking. The resulting complete order is converted into
bounded rank rewards, while boundary expansion, local refinement, and
inactive-anchor pruning adapt the buffer as the policy evolves. Across four
open-ended benchmarks, RankBuffer consistently outperforms all pointwise
baselines. It also achieves nearly on-par performance with the strongest ranking-based reward baseline while substantially reducing judging cost. Ablations demonstrate the importance of both local fine ranking and anchor response content, while buffer analyses show that rollout-derived anchors progressively extend and refine the covered quality scale. These results establish response reuse as an effective approach to efficient relative reward construction. 
\footnote{ Code and reproduction instructions are available at \url{https://github.com/PuffYang/RankBuffer}.
}
\end{abstract}

\section{Introduction}
\label{sec:introduction}

Reinforcement learning (RL) has become an important approach for improving the reasoning and instruction-following capabilities of large language models (LLMs)~\citep{ouyang2022training,shao2024deepseekmath,yu2026dapo}.
Group-based algorithms such as Group Relative Policy Optimization (GRPO)~\citep{shao2024deepseekmath} have emerged as a widely adopted optimization paradigm: instead of relying on a learned value function, GRPO samples a group of rollouts for each query and estimates the advantage of each rollout relative to others within the same group. 
For tasks with verifiable outcomes, such as mathematics and code generation, rewards can often be obtained from exact-match rules, unit tests, or programmatic verifiers~\citep{shao2024deepseekmath,guo2025deepseek}.
In open-ended generation, however, multiple responses may be valid and no unique reference answer is available.
Reward construction for these tasks therefore commonly relies on an LLM judge equipped with query-specific rubrics~\citep{gunjal2026rubrics,huang2025reinforcement,wei2026qurl}.

Most existing rubric-as-reward methods ask the judge to assign a pointwise reward to each rollout independently~\citep{shen2026rethinking,xu2026alternating}. However, such pointwise rewards only provide an indirect and often unstable reward signal for open-ended generation. The challenge is twofold. First, without a canonical answer, the judge must assign pointwise rewards to complex responses on a numerical scale that is difficult to calibrate, and the resulting rewards can vary across queries, rubrics, and training stages~\citep{zheng2023judging,dubois2024length}. Second, GRPO relies on relative differences within a rollout group to estimate advantages, whereas independently assigned pointwise rewards may fail to capture these differences reliably: rewards within a group often cluster on a narrow scale, yielding vanishing or noisy advantages precisely where discriminative signal is most needed.

To address these issues and stabilize group-based RL training, recent studies construct rewards from relative comparisons among rollouts rather than independent pointwise rewards. ArenaRL~\citep{zhang2026arenarl} combines reference-based pre-ranking with a seeded elimination tournament; RRC-AGR and RRC-SCR~\citep{wang2026rrc} derive rewards from pairwise wins against reference responses or among rollouts within the same group; Tournament-GRPO~\citep{yang2026tournament} builds relative rewards through repeated multi-round tournaments. While these methods provide ranking signals that are better suited to group-relative advantage estimation, they incur substantial judging cost: each new rollout group requires multiple comparison or ranking calls, tournament rounds must be executed sequentially, and previously judged responses are discarded rather than reused. As a result, existing ranking-based reward methods trade away the efficiency of pointwise reward methods.

\begin{figure}[htbp]
    \centering
    \includegraphics[width=\linewidth]{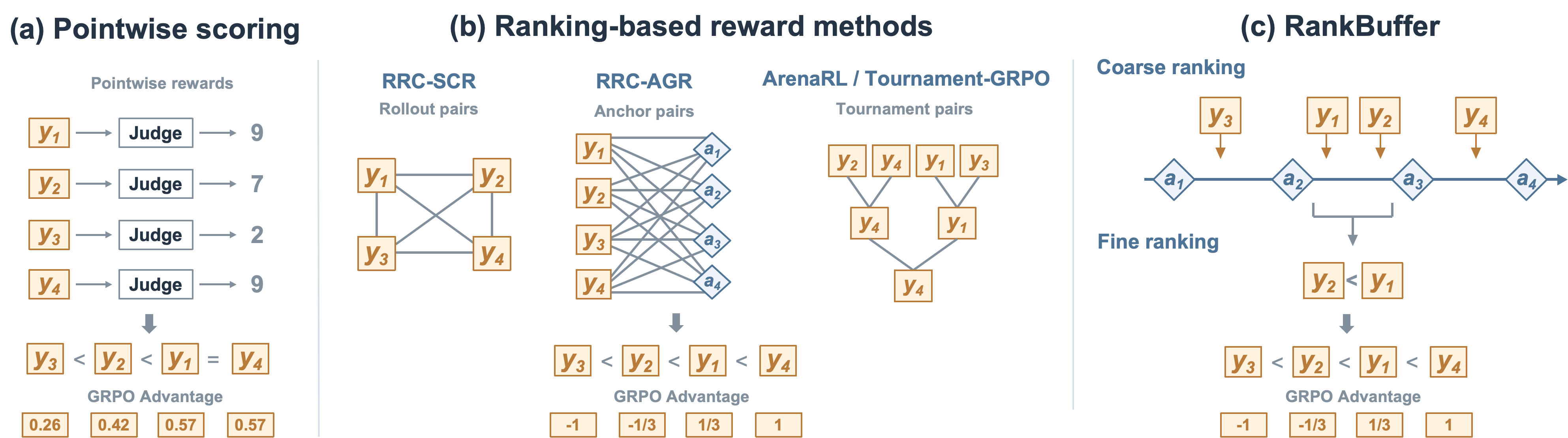}
     \caption{Reward construction with pointwise reward methods, conventional ranking-based reward methods, and RankBuffer. Pointwise rewards are assigned independently, so reward values within a rollout group may cluster or tie and obscure relative differences. Ranking-based reward methods first obtain a complete order and map it to evenly spaced reward levels that preserve the relative quality of the rollouts. RankBuffer produces the same structured reward signal using reusable anchors and selective fine ranking.}
    \label{fig:intro_concept}
\end{figure}

In this work, we ask whether ranking-based reward signals can be obtained at near-pointwise judging cost. We introduce \textbf{RankBuffer}, a buffer-assisted coarse-to-fine ranking framework for open-ended generation. RankBuffer maintains an ordered, query-specific buffer of previously judged responses as a reusable quality scale. Each new rollout is inserted into an anchor interval through a single independent judgment—incurring a per-rollout judging pattern comparable to that of pointwise reward methods and fully parallelizable—and only rollouts assigned to the same interval undergo local fine ranking. Concatenating the local orders yields a complete group ranking, which is converted into bounded rank rewards for GRPO advantage estimation. In this way, RankBuffer amortizes comparison costs across training iterations: judgments made for earlier rollout groups are retained as anchors rather than discarded, so that relative rewards are constructed from accumulated evidence instead of rebuilt from scratch for every group. To keep the buffer aligned with the evolving policy, we further design rule-based maintenance through boundary expansion, local refinement, and inactive-anchor pruning, which progressively extends and refines the covered quality scale.
Figure~\ref{fig:intro_concept} contrasts RankBuffer with pointwise reward methods and conventional ranking-based reward methods.

Our contributions are summarized as follows:

\begin{itemize}
    \item We introduce \textbf{RankBuffer}, a buffer-assisted coarse-to-fine ranking framework for open-ended generation. It reuses an ordered anchor buffer across iterations and combines independent insertion with selective local fine ranking to construct rewards for GRPO.

    \item We design rule-based buffer maintenance through boundary expansion, local refinement, and inactive-anchor pruning. We study Nectar and bootstrap initialization and compare dynamic maintenance with static and LLM-based alternatives.

    \item Across four benchmarks, RankBuffer improves the strongest pointwise baseline by 3.83 average points and achieves comparable performance to other ranking-based reward methods while substantially reducing judging cost. Further analyses examine anchor count, the role of fine ranking, and buffer evolution.
\end{itemize}

\section{Related Work}
\label{sec:related_work}

\subsection{Rubric-Guided Judging and Rewards}

Early LLM evaluators introduced explicit rubrics and rubric-guided
judges~\citep{liu2023g,kim2024prometheus,zhu2025judgelm}. Recent work
develops reasoning reward models, refined rubrics, harder reward-model
evaluation, and localized evidence-based adjudication
~\citep{chen2026rm,shen2026rethinking,malik2026rewardbench,
yang2026auditing}. Nevertheless, LLM judgments remain sensitive to position
and verbosity~\citep{zheng2023judging,dubois2024length}.

Preference-based RL and AI feedback underpin language-model alignment
~\citep{christiano2017deep,stiennon2020learning,ouyang2022training,
bai2022constitutional}. Recent methods extend RL to open-ended tasks with
query-specific rubrics~\citep{gunjal2026rubrics,huang2025reinforcement,
wei2026qurl}. Despite their variety, these methods use pointwise rewards: each rollout is evaluated independently against its rubric and
mapped to a scalar reward. Other work adapts evaluators through learned rubric generation, pairwise comparison, or process supervision
~\citep{xu2026alternating,jia2026open,ding2026evorubrics,
zhang2026oases}. These reward signals can be used with GRPO
~\citep{shao2024deepseekmath} and subsequent group-based objectives
~\citep{liu2024understanding,yu2025dapo,chu2026gpg,liu2026gdpo}.

\subsection{Ranking-Based Reward Methods}

For retrieval, LLMs have been used to rank candidates through pairwise comparisons and other efficient ranking strategies~\citep{qin2024large,chen2025tourrank,liu2025leveraging,liu2025e2rank}. Ranking-based reward methods for RL instead compare policy rollouts through tournaments, reference-guided competition, or pairwise rewards~\citep{yang2026tournament,zhang2026arenarl,wang2026rrc}. RankBuffer differs from these methods by maintaining a persistent ordered response scale with an off-the-shelf judge. Each new rollout is inserted relative to existing anchors to identify its interval, and only rollouts within the same interval require further comparison. This allows RankBuffer to reuse prior judgments without training a specialized reward model or reconstructing comparisons for every rollout group.

\section{Method}
\label{sec:method}

\subsection{Overview}
\label{sec:method_overview}

Figure~\ref{fig:method_overview} summarizes the ranking and buffer-maintenance workflow.
Pseudocode is provided in Appendix~\ref{app:rankbuffer_algorithm}.

Given a query $x$ and its query-specific weighted rubrics
$\mathcal{R}_x$, the policy generates a group of $G$ rollouts:
\begin{equation}
    \mathcal{Y}_x^t
    =
    \{y_1^t,\ldots,y_G^t\},
    \qquad
    y_i^t \sim \pi_{\theta_t}(\cdot \mid x),
\end{equation}
where $t$ denotes the current training iteration and
$\pi_{\theta_t}(\cdot \mid x)$ is the response distribution conditioned on
query $x$ under the current policy with parameters $\theta_t$.

RankBuffer maintains a query-specific buffer of anchor responses:
\begin{equation}
    \mathcal{A}_x^t
    =
    (a_1^t,\ldots,a_{K_t}^t),
    \qquad
    a_1^t \prec \cdots \prec a_{K_t}^t,
\end{equation}
where $a_k^t$ is the $k$-th anchor response stored for query $x$ at
iteration $t$, $K_t$ is the number of anchors in this buffer, and $\prec$
denotes the stored quality order from worst to best.

RankBuffer performs coarse ranking followed by selective fine ranking,
maps the resulting group order to rewards, and updates the buffer for
subsequent training iterations.

\begin{figure}[t]
    \centering
    \includegraphics[width=0.9\linewidth]{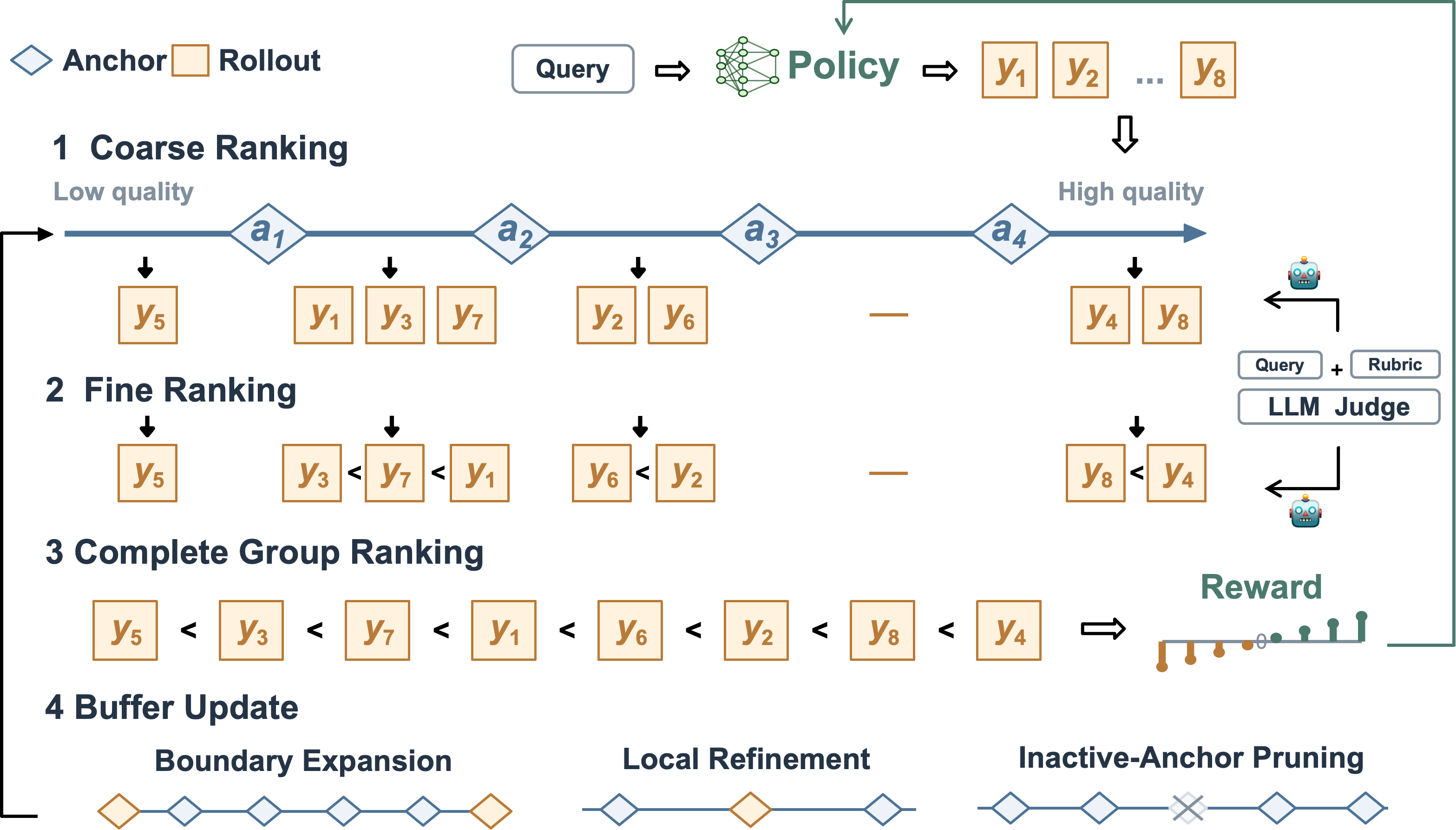}
    \caption{RankBuffer workflow. An ordered, query-specific buffer enables
parallel coarse ranking, followed by local fine ranking within intervals
containing multiple rollouts. The resulting complete ranking determines
the rewards and GRPO advantages. Boundary expansion, local refinement,
and inactive-anchor pruning update the buffer for subsequent iterations.}
    \label{fig:method_overview}
\end{figure}

\subsection{Buffer Initialization}
\label{sec:buffer_initialization}

We consider two ways to construct the initial query-specific buffer.

\paragraph{Nectar initialization.}

For \textbf{RankBuffer}, we select $K_0$ responses from
the ranked responses associated with query $x$ in
Nectar~\citep{zhu2024starling}. The selected anchors are stored from worst
to best:
\begin{equation}
    \mathcal{A}_x^0
    =
    (a_1^0,\ldots,a_{K_0}^0),
\end{equation}
where $K_0$ denotes the initial buffer size.

\paragraph{Bootstrap initialization.}

For \textbf{RankBuffer-Bootstrap}, we use listwise judging during the first
training epoch to rank the initial rollout group for each query $x$ from
worst to best:
\begin{equation}
    \sigma_x^0
    =
    J
    \left(
        x,
        \mathcal{R}_x,
        \mathcal{Y}_x^0
    \right),
\end{equation}
where $J$ denotes the LLM judge and $\sigma_x^0$ is the resulting
worst-to-best rollout ordering. We then select
$K_0$ rollouts from this ordering as anchors to initialize the buffer:
\begin{equation}
    \mathcal{A}_x^0
    =
    \operatorname{Select}_{K_0}
    \left(
        \sigma_x^0
    \right).
\end{equation}

\subsection{Buffer-Assisted Coarse-to-Fine Ranking}
\label{sec:coarse_to_fine}

\paragraph{Coarse ranking.}

The $K_t$ ordered anchors divide the response-quality space into
$K_t+1$ intervals. For each rollout $y_i^t$, the judge predicts its
insertion position:
\begin{equation}
    p_i^t
    =
    J
    \left(
        x,
        \mathcal{R}_x,
        \mathcal{A}_x^t,
        y_i^t
    \right)
    \in
    \{0,\ldots,K_t\}.
\end{equation}

Position $p_i^t=0$ places the rollout below the lowest anchor, while
$p_i^t=K_t$ places it above the highest anchor. An internal position
$p_i^t$ places the rollout between $a_{p_i^t}^t$ and $a_{p_i^t+1}^t$.

The insertion positions partition the rollout group into buckets:
\begin{equation}
    \mathcal{C}_p^t
    =
    \{y_i^t \mid p_i^t=p\},
    \qquad
    p\in\{0,\ldots,K_t\}.
\end{equation}

Here, $\mathcal{C}_p^t$ contains all rollouts with insertion position $p$
at iteration $t$.

All insertion judgments are performed independently and can be executed
in parallel.

\paragraph{Fine ranking.}

For every bucket containing at least two rollouts, i.e.,
$|\mathcal{C}_p^t|\geq 2$, we perform listwise ranking within the bucket.
The judge returns a permutation of the candidate IDs
from worst to best:
\begin{equation}
    \sigma_p^t
    =
    J
    \left(
        x,
        \mathcal{R}_x,
        \mathcal{C}_p^t
    \right).
\end{equation}

Here, $\sigma_p^t$ denotes the local rollout order for bucket $p$.
The final group ordering first follows the insertion positions $p_i^t$
in ascending order, with rollouts sharing a position ordered by the
fine-ranking result. Single-rollout buckets require no fine ranking.
We denote the resulting complete worst-to-best ordering by $\sigma^t$.
The coarse- and fine-ranking prompts are provided in
Appendix~\ref{app:ranking_prompts}.

\subsection{Rank-Based Reward and Advantage Estimation}
\label{sec:rank_reward}

Let $q_i^t\in\{0,\ldots,G-1\}$ denote the position of rollout $y_i^t$
in the final worst-to-best ordering $\sigma^t$. We map this position
linearly to a rank reward:
\begin{equation}
    r_{i,t}^{\mathrm{rank}}
    =
    -1
    +
    \frac{2q_i^t}{G-1}.
\end{equation}

The training reward additionally includes an auxiliary format reward:
\begin{equation}
    r_{i,t}
    =
    r_{i,t}^{\mathrm{rank}}
    +
    \lambda_{\mathrm{fmt}}
    r_{i,t}^{\mathrm{fmt}},
\end{equation}
where $r_{i,t}^{\mathrm{fmt}}$ indicates whether the rollout follows the
required output format.

GRPO normalizes the rewards within each rollout group to estimate the
advantages:
\begin{equation}
    \widehat{A}_{i,t}
    =
    \frac{
        r_{i,t}-\overline{r}_t
    }{
        \sqrt{
            \frac{1}{G-1}
            \sum_{j=1}^{G}
            \left(r_{j,t}-\overline{r}_t\right)^2
        }
        +\epsilon
    },
\end{equation}
where
\begin{equation}
    \overline{r}_t
    =
    \frac{1}{G}
    \sum_{j=1}^{G}r_{j,t}.
\end{equation}

The estimated advantages are then used in the standard GRPO policy
objective.

\subsection{Rule-Based Buffer Update}
\label{sec:buffer_update}

After ranking the current rollout group, we update the query-specific
buffer using boundary expansion, local refinement, and inactive-anchor
pruning.

\paragraph{Boundary expansion.}

If one or more rollouts are assigned below the lowest anchor, we add the
lowest-ranked rollout in $\mathcal{C}_0^t$ to the beginning of the
buffer. Similarly, if one or more rollouts are assigned above the
highest anchor, we add the highest-ranked rollout in
$\mathcal{C}_{K_t}^t$ to the end of the buffer. At most one response is
added at each boundary for a query in one epoch.

\paragraph{Local refinement.}

For an internal bucket $\mathcal{C}_p^t$, where
$p\in\{1,\ldots,K_t-1\}$, if $|\mathcal{C}_p^t|\geq 3$, we add its
lower-median rollout as a new anchor:
\begin{equation}
    a_{\mathrm{new}}
    =
    \sigma_p^t
    \left[
        \left\lceil
        \frac{|\mathcal{C}_p^t|}{2}
        \right\rceil
    \right],
    \qquad
    |\mathcal{C}_p^t|\geq 3.
\end{equation}

The new anchor is inserted between the two anchors defining that
interval.

\paragraph{Inactive-anchor pruning.}

For each internal anchor $a_k^t$, where $k\in\{2,\ldots,K_t-1\}$,
we track whether either of its adjacent intervals receives a rollout.
The inactivity counter $h_k$ is initialized to zero when the anchor is
created and updated once per epoch as
\begin{equation}
    h_k^t
    =
    \begin{cases}
        0,
        &
        \mathcal{C}_{k-1}^t
        \cup
        \mathcal{C}_k^t
        \neq \varnothing,
        \\[3pt]
        h_k^{t-1}+1,
        &
        \text{otherwise}.
    \end{cases}
\end{equation}

An internal anchor is removed after being inactive for $H$ consecutive
epochs. The lowest and highest anchors
are never pruned. The resulting ordered buffer becomes
$\mathcal{A}_x^{t+1}$ and is used to rank future rollout groups.

\section{Experiments}
\label{sec:experiments}

\subsection{Experimental Setup}
\label{sec:experimental_setup}

\paragraph{Training framework and models.}
We build our RL pipeline on top of verl\footnote{\url{https://github.com/verl-project/verl}} and optimize the policy with GRPO~\citep{shao2024deepseekmath}.
The policy is Qwen3-8B~\citep{yang2025qwen3}, and the LLM judge is a locally deployed Qwen3-30B-A3B-Instruct-2507~\citep{yang2025qwen3}.
Unless specified otherwise, all methods use the same policy, judge, and number of training steps.
We sample $G=8$ rollouts per query and report results after four training
epochs (200 steps). Detailed hyperparameters and hardware settings are
provided in Appendix~\ref{app:implementation_details}.

\paragraph{Training data and buffer initialization.}
Following the two initialization strategies in
Section~\ref{sec:buffer_initialization}, Nectar initialization uses 1,000
queries randomly sampled from
Nectar~\citep{zhu2024starling} with seed 42, selecting four initial anchors
per query at Nectar ranks $(7,5,3,1)$.
Bootstrap initialization uses the same 1,000 queries but constructs its
anchors from policy rollouts instead of Nectar responses: during the first
epoch, each query's initial rollout group is ranked listwise using the
prompt in Appendix~\ref{app:listwise_prompt}, and four
rollouts at worst-to-best positions $(1,3,6,8)$ are selected as anchors.

\paragraph{Query-specific rubrics.}
Each query is paired with three to five weighted, query-specific rubrics generated once by DeepSeek-V4-Flash~\citep{xu2026deepseek} and then held fixed.
Rubric-generation details and prompts are provided in Appendix~\ref{app:rubric_generation}.

\paragraph{Evaluation.}
We evaluate on AlpacaEval~2~\citep{dubois2024length}, Arena-Hard~v2~\citep{li2024crowdsourced}, WildBench~v2~\citep{lin2025wildbench}, and WritingBench~\citep{wu2026writingbench}.
We use Qwen3-30B-A3B-Instruct-2507~\citep{yang2025qwen3} as the judge for all four benchmarks.

\subsection{Reward Quality and Judging Efficiency}
\label{sec:baselines}
\label{sec:pointwise_results}
\label{sec:efficiency_results}

\textbf{RankBuffer} uses Nectar initialization (Section~\ref{sec:buffer_initialization})
with rule-based updates (Section~\ref{sec:buffer_update}). \textbf{RankBuffer-Bootstrap} uses bootstrap initialization
(Section~\ref{sec:buffer_initialization}) and the same rule-based buffer
updates as RankBuffer.
Implementation details are provided in Appendix~\ref{app:baseline_details}.
We compare RankBuffer with two categories of baselines:

\paragraph{Pointwise reward methods.}
These baselines assign pointwise rewards to rollouts independently using
the query-specific rubrics. We include GRPO, GDPO~\citep{liu2026gdpo},
Dr.\ GRPO~\citep{liu2025understanding}, DAPO~\citep{yu2025dapo}, and
GPG~\citep{chu2026gpg}.
Table~\ref{tab:pointwise_main} reports the results.
RankBuffer outperforms all pointwise baselines on all four benchmarks.

\begin{table}[htbp]
\centering
\begin{minipage}[c]{0.48 \linewidth}
\caption{Comparison with pointwise baselines.}
\label{tab:pointwise_main}
\begin{center}
\begin{tabular}{lrrrr}
\toprule
Method & AE2 & \shortstack{AH-v2} & \shortstack{WB-v2} & WrB \\
\midrule
Base             & 31.11 &  9.30 & 37.23 & 62.49 \\
GRPO             & 72.73 & \underline{27.50} & 46.57 & 65.42 \\
GDPO             & 64.35 & 20.10 & 45.14 & 64.94 \\
Dr.\ GRPO        & 56.15 & 18.90 & 44.32 & 63.37 \\
DAPO             & \underline{74.22} & 26.60 & \underline{46.83} & \underline{66.13} \\
GPG              & 53.04 & 16.90 & 45.78 & 64.98 \\
\midrule
\textbf{RankBuffer} & \textbf{82.80} & \textbf{30.20} & \textbf{48.79} & \textbf{67.27} \\
\bottomrule
\end{tabular}
\end{center}
\end{minipage}
\hfill
\begin{minipage}[c]{0.45\linewidth}
    \centering
    \includegraphics[width=\linewidth]{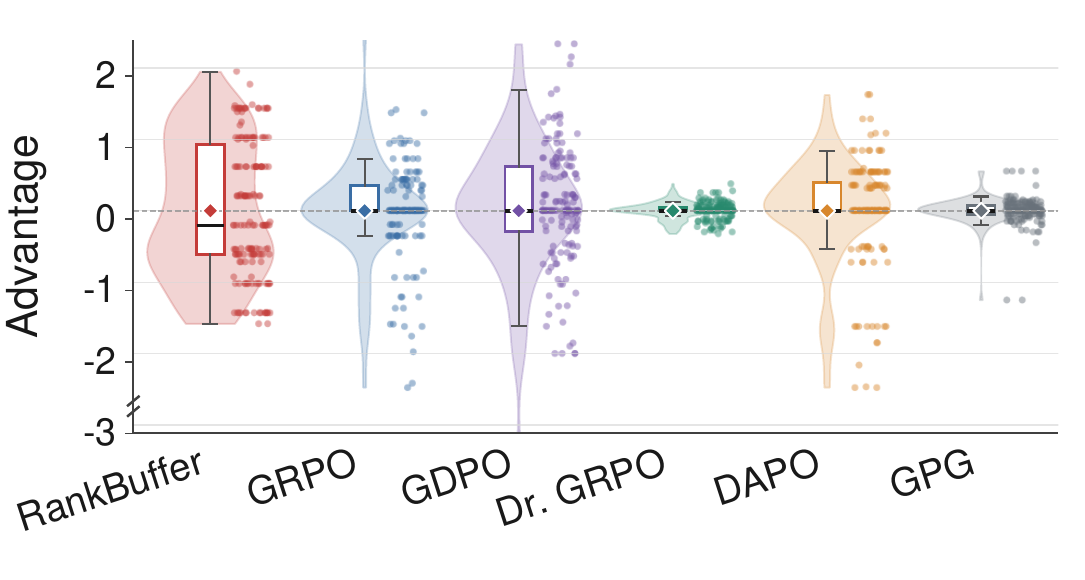}
    \captionof{figure}{Advantages at step 191.}
    \label{fig:advantage_distribution}
\end{minipage}
\end{table}

Figure~\ref{fig:advantage_distribution} shows the normalized advantages at training step 191.
RankBuffer converts rollout rankings into distinct advantage levels, providing a richer signal for within-group optimization.
In contrast, pointwise reward baselines often produce clustered advantages, offering less differentiation among rollouts.

\begin{figure}[htbp]
    \centering
    \begin{minipage}[t]{0.485\linewidth}
        \centering
        \includegraphics[width=\linewidth]{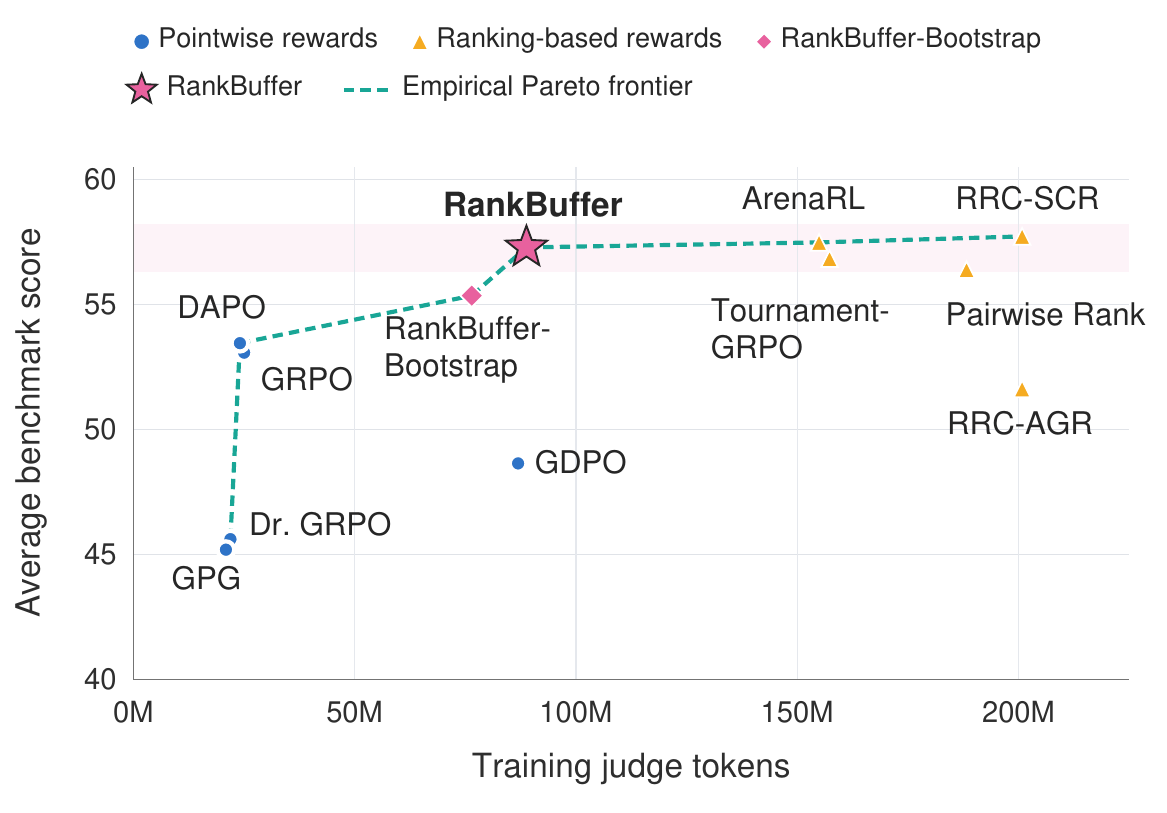}
        \caption{Average score versus judge tokens.}
        \label{fig:reward_efficiency}
    \end{minipage}
    \hfill
    \begin{minipage}[t]{0.485\linewidth}
        \centering
        \includegraphics[width=\linewidth]{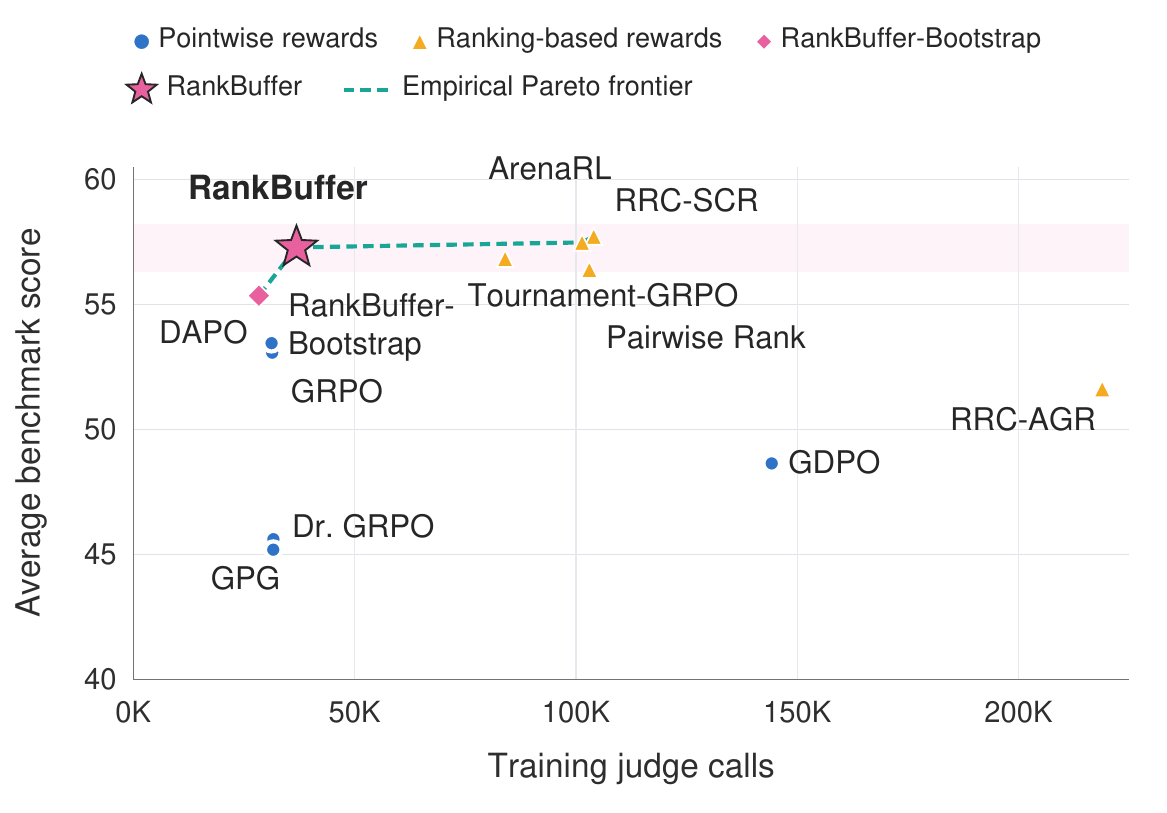}
        \caption{Average score versus judge calls.}
        \label{fig:reward_efficiency_calls}
    \end{minipage}
\end{figure}

\paragraph{Ranking-based reward methods.}
We include
Tournament-GRPO~\citep{yang2026tournament}, RRC-AGR and
RRC-SCR~\citep{wang2026rrc}, Pairwise Rank
(exhaustive pairwise judging followed by win-rate ranking; Appendix~\ref{app:pairwise_rank}), and
ArenaRL~\citep{zhang2026arenarl}.
These baselines construct rewards through relative comparisons rather
than pointwise rewards.
Figures~\ref{fig:reward_efficiency} and~\ref{fig:reward_efficiency_calls}
compare average scores against total judge input and output tokens and
successfully parsed judge requests during RL training, respectively.
RankBuffer achieves comparable or better performance than the ranking-based reward baselines while substantially reducing judge-token cost.
It achieves nearly the same performance as the strongest ranking-based reward baseline, RRC-SCR, while using less than half its judge tokens.
Full results are provided in Appendix~\ref{app:full_results}.

\subsection{Buffer Design Analysis}
\label{sec:buffer_analysis}

\paragraph{Initialization and dynamic maintenance.}
RankBuffer* and RankBuffer-Bootstrap* freeze their buffers at the end of
the first epoch and perform no further buffer updates.
Table~\ref{tab:buffer_variants} compares these four configurations.
RankBuffer achieves the highest average score ($57.27$), outperforming
RankBuffer* by $2.26$ points and demonstrating the benefit of adapting
Nectar-initialized anchors during training.
For bootstrap initialization, RankBuffer-Bootstrap* yields a higher average
score and lower judge cost than RankBuffer-Bootstrap, suggesting that
policy-generated initial anchors can also provide a useful fixed reference.

\begin{table}[htbp]
\caption{Effect of buffer initialization and maintenance. An asterisk (*) denotes static buffers.}
\label{tab:buffer_variants}
\begin{center}
\begin{tabular}{lrrrrrrr}
\toprule
Method & AE2 & \shortstack{AH-v2} & \shortstack{WB-v2} & Writing & Avg. & \shortstack{Tokens(M)} & \shortstack{Calls(K)} \\
\midrule
RankBuffer & 82.80 & 30.20 & 48.79 & 67.27 & \textbf{57.27} & 88.8M & 36.9K \\
RankBuffer* & 79.69 & 26.70 & 46.95 & 66.71 & 55.01 & 68.0M & 36.2K \\
\midrule
RankBuffer-Bootstrap & 79.50 & 28.40 & 47.31 & 66.15 & 55.34 & 76.5M & 28.4K \\
RankBuffer-Bootstrap* & 81.61 & 29.60 & 47.47 & 67.48 & \textbf{56.54} & 60.7M & 28.1K \\
\bottomrule
\end{tabular}
\end{center}
\end{table}

\paragraph{LLM buffer variants.}
We additionally test whether the buffer-update rules can be replaced by LLM.
After each ranking step, LLM receives the latest combined worst-to-best
ordering of the current anchors and newly ranked rollouts, and selects
four responses from this ordering as the next-step anchors.
A second variant, LLM + look-ahead, additionally provides LLM with the
next epoch's unranked rollouts as context.
The detailed update procedures and prompts are provided in
Appendix~\ref{app:llm_buffer}.

As shown in Table~\ref{tab:llm_buffer}, LLM reaches an average score of $56.60$, which is $0.67$ points below RankBuffer, while using $8.7\%$ more judge tokens and $12.8\%$ more calls.
LLM + look-ahead achieves a lower average score ($54.98$) than LLM ($56.60$),
indicating that providing the next epoch's unranked rollouts as context for
buffer updates does not improve downstream performance in this setting.

\begin{table}[htbp]
\caption{Comparison of RankBuffer, RankBuffer-Bootstrap, and LLM-based buffer management.}
\label{tab:llm_buffer}
\begin{center}
\begin{tabular}{lrrrrrrr}
\toprule
Method & AE2 & \shortstack{AH-v2} & \shortstack{WB-v2} & Writing & Avg. & \shortstack{Tokens(M)} & \shortstack{Calls(K)} \\
\midrule
RankBuffer & \textbf{82.80} & 30.20 & \textbf{48.79} & \textbf{67.27} & \textbf{57.27} & 88.8M & 36.9K \\
RankBuffer-Bootstrap & 79.50 & 28.40 & 47.31 & 66.15 & 55.34 & 76.5M & 28.4K \\
LLM & 81.74 & \textbf{30.50} & 47.34 & 66.83 & 56.60 & 96.5M & 41.6K \\
LLM + look-ahead & 79.07 & 28.40 & 46.31 & 66.13 & 54.98 & 111.0M & 42.1K \\
\bottomrule
\end{tabular}
\end{center}
\end{table}

\subsection{How Does the Dynamic Buffer Evolve?}
\label{sec:buffer_evolution}

\begin{figure}[htbp]
    \centering
    \begin{minipage}[t]{0.485\linewidth}
        \centering
        \textbf{(a)}\\
        \includegraphics[width=\linewidth]{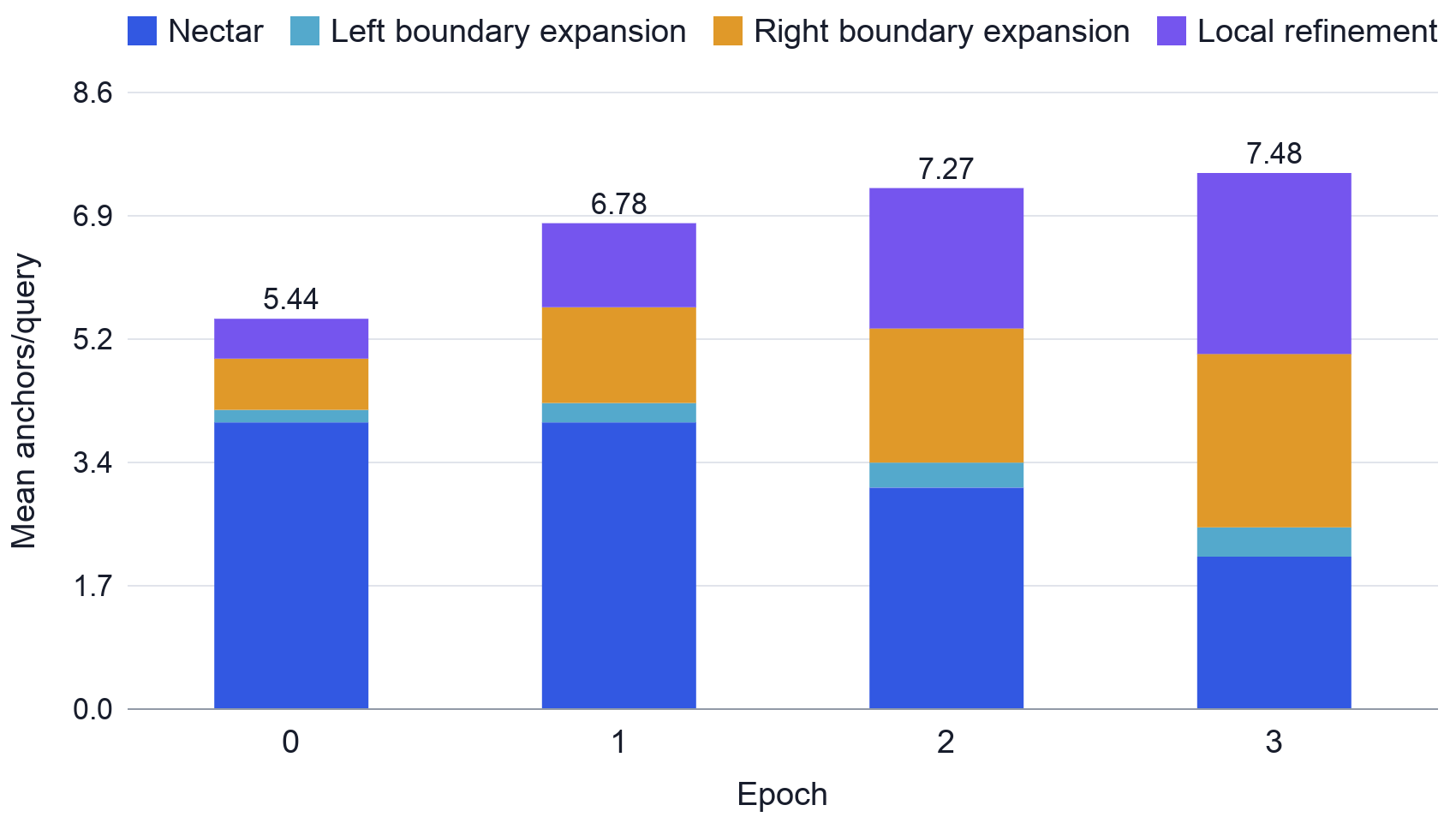}
    \end{minipage}
    \hfill
    \begin{minipage}[t]{0.485\linewidth}
        \centering
        \textbf{(b)}\\
        \includegraphics[width=\linewidth]{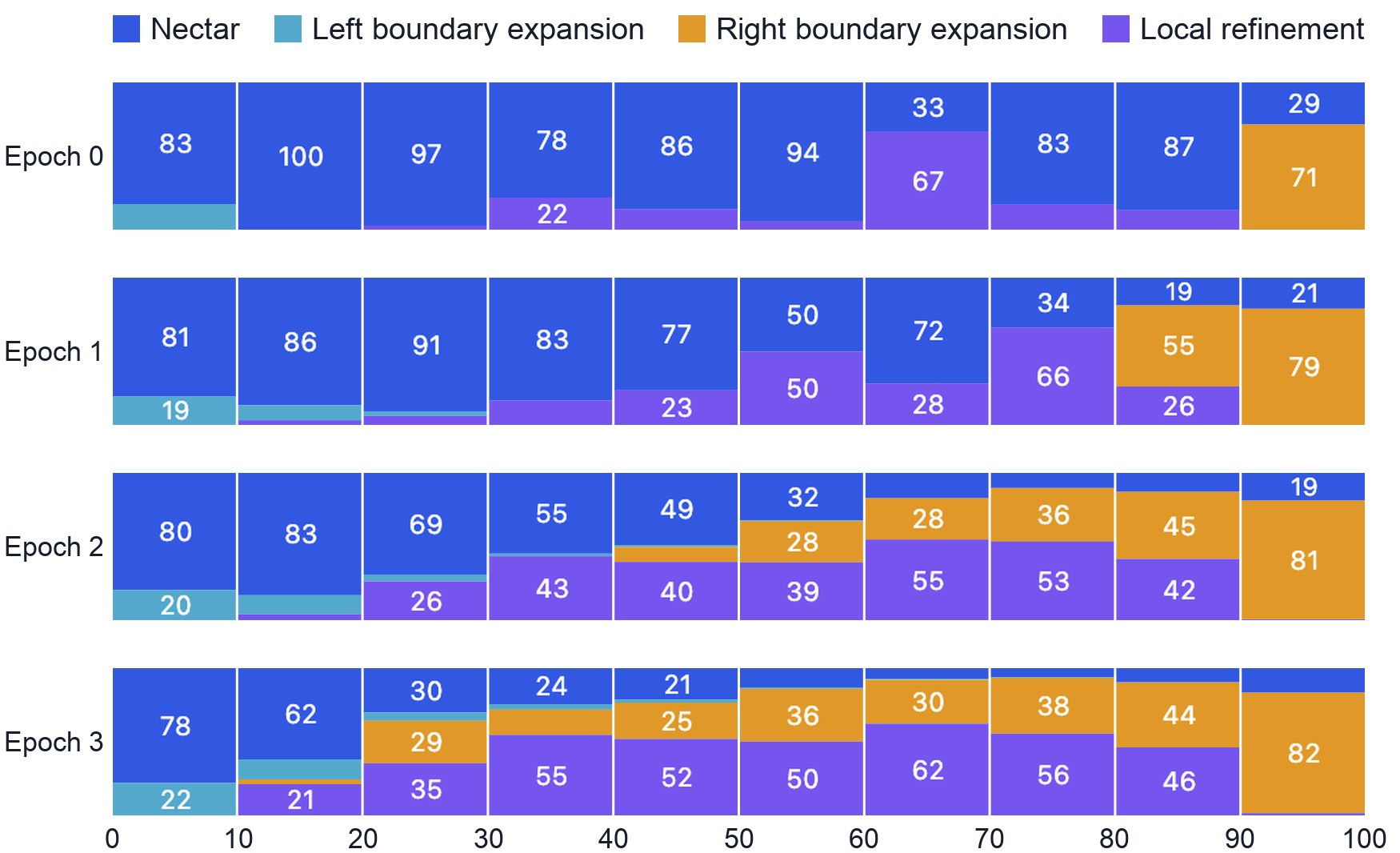}
    \end{minipage}
    \caption{Dynamic buffer evolution: \textbf{(a)} anchor counts by source; \textbf{(b)} source composition along the quality scale.}
    \label{fig:buffer_evolution}
\end{figure}

Figure~\ref{fig:buffer_evolution}(a) tracks the source of every live anchor at the end of each epoch.
Starting from four Nectar anchors per query, the mean buffer size grows to 5.44 after epoch~0 and then to 6.78, 7.27, and 7.48.
The mean number of Nectar anchors declines, while right-boundary expansion
and local-refinement anchors increase substantially. Left-boundary expansion
anchors also increase, but remain a small component of the buffer.
The buffer thus progressively incorporates policy-generated anchors,
primarily through upper-boundary expansion and interior refinement.
Growth slows over time as the removal of inactive anchors offsets the
insertion of new rollout anchors.
The buffer thus adapts to policy changes while exhibiting progressively
smaller size increases over the observed training horizon.

Figure~\ref{fig:buffer_evolution}(b) shows anchor sources along the quality
scale. To compare different-length buffers, we normalize anchor ranks to
0--100 and plot source proportions in ten bins at each epoch end.
Initial Nectar anchors shift toward lower normalized positions, while local-refinement
anchors concentrate in higher-quality interior regions. Right-boundary
expansion anchors dominate the highest-quality bin, increasing from
$71\%$ in epoch~0 to $82\%$ in epoch~3. This trend is consistent with
progressive improvements in overall rollout quality during training,
with responses extending beyond the original buffer's upper quality boundary.
Together, these analyses show that buffer updates extend the covered quality
range and improve ranking resolution where current rollouts concentrate.

\subsection{Ablation Studies}
\label{sec:ablations}

\paragraph{Number of initial anchors.}
Table~\ref{tab:anchor_count_ablation} varies the initial buffer size
$K_0\in\{2,4,7\}$ for both Nectar and bootstrap initialization.
With Nectar initialization, $K_0=4$
performs best, improving the average by $1.17$ points over $K_0=2$ and by
$0.87$ points over $K_0=7$. The seven-anchor setting also consumes $28.4\%$
more judge tokens than the four-anchor setting. With bootstrap initialization,
the best result instead occurs at $K_0=7$, which reaches $57.24$ but uses
$20.8\%$ more judge tokens than $K_0=4$. Thus, increasing the number of
anchors is not uniformly beneficial. The exact anchor subsets are
specified in Appendix~\ref{app:ablation_details}.

\begin{table}[htbp]
\caption{Effect of the initial anchor count.}
\label{tab:anchor_count_ablation}
\begin{center}
\begin{tabular}{lcrrrrrrr}
\toprule
Init. & $K_0$ & AE2 & AH-v2 & WB-v2 & Writing & Avg. & Tokens (M) & Calls (K) \\
\midrule
\multirow{3}{*}{Nectar} & 2 & 80.99 & 30.00 & 46.73 & 66.69 & 56.10 & 79.3M & 36.4K \\
         & 4 & 82.80 & 30.20 & 48.79 & 67.27 & \textbf{57.27} & 88.8M & 36.9K \\
         & 7 & 81.06 & 29.60 & 48.02 & 66.92 & 56.40 & 114.1M & 36.8K \\
\midrule
\multirow{3}{*}{Bootstrap} & 2 & 80.62 & 27.90 & 47.55 & 66.79 & 55.72 & 66.5M & 28.0K \\
         & 4 & 79.50 & 28.40 & 47.31 & 66.15 & 55.34 & 76.5M & 28.4K \\
         & 7 & 79.50 & 33.50 & 48.58 & 67.39 & \textbf{57.24} & 92.4M & 28.7K \\
\bottomrule
\end{tabular}
\end{center}
\end{table}

\paragraph{Removing local fine ranking.}
Starting from RankBuffer, we evaluate a \emph{coarse-only} variant that
performs coarse ranking but omits local fine ranking;
rollouts assigned to the same interval receive the same average rank.
Table~\ref{tab:coarse_only_ablation} shows that this variant reduces judge-token
use by $16.5\%$ and judge calls by $15.9\%$, but lowers all four benchmark
scores and reduces the average from $57.27$ to $54.34$.

\begin{table}[htbp]
\caption{Coarse-only ablation.}
\label{tab:coarse_only_ablation}
\begin{center}
\begin{tabular}{lrrrrrrr}
\toprule
Variant & AE2 & AH-v2 & WB-v2 & Writing & Avg. & Tokens (M) & Calls (K) \\
\midrule
RankBuffer & \textbf{82.80} & \textbf{30.20} & \textbf{48.79} & \textbf{67.27} & \textbf{57.27} & 88.8M & 36.9K \\
Coarse-only & 78.30 & 27.60 & 45.56 & 65.89 & 54.34 & 74.1M & 31.0K \\
\bottomrule
\end{tabular}
\end{center}
\end{table}

\paragraph{Removing anchor content.}
We additionally evaluate an \emph{order-only} variant in which the coarse
judge sees only the ordered anchor identifiers, rather than their response
content, while local fine ranking remains unchanged. As shown in
Table~\ref{tab:order_only_ablation}, AlpacaEval~2 drops from $82.80$ to
$51.93$ with Nectar initialization and from $79.50$ to $54.53$ with
bootstrap initialization. Thus, the anchor order alone is insufficient for
placing new responses reliably.

\begin{table}[htbp]
\caption{Order-only ablation.}
\label{tab:order_only_ablation}
\begin{center}
\begin{tabular}{lcr}
\toprule
Initialization & Variant & AlpacaEval~2 \\
\midrule
\multirow{2}{*}{Nectar} & RankBuffer & \textbf{82.80} \\
                        & Order-only & 51.93 \\
\multirow{2}{*}{Bootstrap} & RankBuffer & \textbf{79.50} \\
                        & Order-only & 54.53 \\
\bottomrule
\end{tabular}
\end{center}
\end{table}

\section{Conclusion}
\label{sec:conclusion}

We introduced RankBuffer, a buffer-assisted framework for constructing
relative rewards in open-ended group-based RL. RankBuffer reuses previously
judged responses as an ordered, query-specific quality scale, combining
parallel coarse ranking with selective local fine ranking to obtain a
complete rollout order. Boundary expansion, local refinement, and
inactive-anchor pruning adapt this scale as the policy evolves. Across four benchmarks, RankBuffer consistently outperforms all pointwise
baselines and achieves performance comparable to the strongest ranking-based
reward methods, while substantially reducing judging cost. The ablations demonstrate that local fine ranking improves
downstream performance and that anchor response content is necessary for
reliable coarse ranking. Buffer analyses further show that rollout-derived
anchors progressively extend the covered quality range and increase ranking
resolution in regions where current rollouts concentrate. Overall, RankBuffer
provides an effective and efficient approach to relative reward construction
by reusing prior judgments without requiring a separately trained reward
model.

\subsection*{AI use statement}

We used generative AI tools only to aid and polish the writing of this paper,
including improving grammar, clarity, and presentation. All AI-assisted edits
were reviewed and revised by the authors, who take full responsibility for the
final content.

\subsection*{Ethics statement}

Our study relies exclusively on publicly accessible datasets and involves no collection of personally identifiable information. Nevertheless, policies optimized with feedback from LLM judges may inherit limitations of the judge models, including systematic biases. The resulting models may also generate inaccurate information, unsubstantiated statements, or biased responses. These risks should be carefully considered when deploying the proposed framework in practical scenarios, particularly in high-stakes domains where such errors may have harmful consequences. For these applications, we encourage additional safeguards, such as factuality assessment, source and citation validation, and appropriate human review.

\subsection*{Reproducibility statement}

The main text describes the RankBuffer method, experimental setup, and
evaluation protocol. The appendix provides complete pseudocode, judging
prompts, implementation settings, baseline details, and ablation
configurations. We also report the complete numerical results to facilitate
verification and reproduction of our experiments. An anonymous repository
containing the code and reproduction instructions is linked in the abstract.

\bibliography{iclr2027_conference}
\bibliographystyle{iclr2027_conference}

\appendix

\section{RankBuffer Pseudocode}
\label{app:rankbuffer_algorithm}

Algorithm~\ref{alg:rankbuffer} constructs rewards for one query group;
Algorithm~\ref{alg:rankbuffer_update} maintains its ordered buffer.
At each training step, Algorithm~\ref{alg:rankbuffer} is applied to each
query in the batch, and the resulting advantages are used in the standard
GRPO policy update. Each anchor's inactivity counter is initialized to zero
when the anchor is created. Buffers and maintenance counters persist throughout training.

\begin{algorithm}[htbp]
\caption{RankBuffer: group ranking and reward construction}
\label{alg:rankbuffer}
\begin{algorithmic}[1]
\small
\Require Query $x$, rubrics $\mathcal{R}_x$, rollouts $\mathcal{Y}_x^t$ of size $G$, epoch $e$
\Require Persistent buffer $\mathcal{A}_x$, initialization source, $K_0$, judge $J$
\Require Format coefficient $\lambda_{\mathrm{fmt}}$, normalization constant $\epsilon$, pruning threshold $H$
\Ensure Rewards $r_i$, advantages $\widehat{A}_i$, updated buffer $\mathcal{A}_x$
\If{$\mathcal{A}_x$ is uninitialized and the source is Nectar}
    \State $\mathcal{A}_x \gets \operatorname{Select}_{K_0}(\text{Nectar responses for }x\text{, worst to best})$
    \State Initialize anchor inactivity counters and per-query epoch markers
\EndIf
\If{$\mathcal{A}_x$ is uninitialized} \Comment{Bootstrap: first-epoch query group}
    \State $\sigma \gets J(x,\mathcal{R}_x,\mathcal{Y}_x^t)$ \Comment{Listwise, worst to best}
    \State $\mathcal{A}_x \gets \operatorname{Select}_{K_0}(\sigma)$
    \State Initialize anchor inactivity counters and per-query epoch markers
\Else
    \State $K \gets |\mathcal{A}_x|$; initialize empty buckets $\mathcal{C}_0,\ldots,\mathcal{C}_K$
    \ForAll{$y_i^t\in\mathcal{Y}_x^t$ in parallel}
        \State $p_i \gets J(x,\mathcal{R}_x,\mathcal{A}_x,y_i^t)\in\{0,\ldots,K\}$
        \State Assign candidate $i$ to $\mathcal{C}_{p_i}$
    \EndFor
    \For{$p=0,\ldots,K$}
        \If{$|\mathcal{C}_p|\geq 2$}
            \State $\sigma_p \gets J(x,\mathcal{R}_x,\mathcal{C}_p)$ \Comment{Listwise, worst to best}
        \Else
            \State $\sigma_p \gets \mathcal{C}_p$ \Comment{Empty or singleton: no judge call}
        \EndIf
    \EndFor
    \State $\sigma \gets \sigma_0\mathbin\Vert\cdots\mathbin\Vert\sigma_K$ \Comment{Concatenate in interval order}
    \State $\mathcal{A}_x \gets \Call{UpdateBuffer}{\mathcal{A}_x,\{\sigma_p\}_{p=0}^K,e,H}$
\EndIf
\For{$i=1,\ldots,G$}
    \State $q_i \gets \text{zero-based position of candidate }i\text{ in }\sigma$
    \State $r_i \gets -1+2q_i/(G-1)+\lambda_{\mathrm{fmt}}r_i^{\mathrm{fmt}}$
\EndFor
\State $\bar r \gets G^{-1}\sum_{i=1}^G r_i$;
      $s \gets \sqrt{(G-1)^{-1}\sum_{i=1}^G(r_i-\bar r)^2}$
\State $\widehat{A}_i \gets (r_i-\bar r)/(s+\epsilon)$ for $i=1,\ldots,G$
\State \Return $\{r_i,\widehat{A}_i\}_{i=1}^G$, $\mathcal{A}_x$
\end{algorithmic}
\end{algorithm}

Here, $\operatorname{Select}_{K_0}$ uniformly subsamples an ordered sequence
while retaining both endpoints; the experimental subsets are listed in
Appendix~\ref{app:ablation_details}. Bootstrap initialization uses the
first group ranking both to compute rewards and to select anchors, without
an additional maintenance update on that group. In Algorithm~\ref{alg:rankbuffer_update},
temporary coordinates refer to the pre-update buffer; additions and pruning
are merged simultaneously so that pruning does not shift insertion intervals.
New anchors have inactivity counter zero.

\begin{algorithm}[t]
\caption{Rule-based buffer maintenance}
\label{alg:rankbuffer_update}
\begin{algorithmic}[1]
\small
\Function{UpdateBuffer}{$\mathcal{A}_x=(a_1,\ldots,a_K),\{\sigma_p\}_{p=0}^K,e,H$}
    \State $P\gets\varnothing$ \Comment{Indices of anchors to prune}
    \If{inactivity has not yet been observed for query $x$ in epoch $e$}
        \For{$k=2,\ldots,K-1$}
            \If{$|\sigma_{k-1}|+|\sigma_k|>0$}
                \State $h_k\gets 0$
            \Else
                \State $h_k\gets h_k+1$
            \EndIf
            \If{$h_k\geq H$}
                \State $P\gets P\cup\{k\}$
            \EndIf
        \EndFor
        \State Mark inactivity as observed for query $x$ in epoch $e$
    \EndIf
    \State $M\gets\{(k,a_k):1\leq k\leq K,\ k\notin P\}$
    \If{$|\sigma_0|>0$ and the left boundary has not expanded in epoch $e$}
        \State Add $(\tfrac12,\sigma_0[1])$ to $M$ \Comment{Worst rollout below the buffer}
        \State Mark the left boundary as expanded in epoch $e$
    \EndIf
    \If{$|\sigma_K|>0$ and the right boundary has not expanded in epoch $e$}
        \State Add $(K+\tfrac12,\sigma_K[|\sigma_K|])$ to $M$ \Comment{Best rollout above the buffer}
        \State Mark the right boundary as expanded in epoch $e$
    \EndIf
    \For{$p=1,\ldots,K-1$}
        \If{$|\sigma_p|\geq 3$}
            \State $j\gets\lceil|\sigma_p|/2\rceil$ \Comment{One-based lower median}
            \State Add $(p+\tfrac12,\sigma_p[j])$ to $M$
        \EndIf
    \EndFor
    \State Initialize $h=0$ for every newly added anchor
    \State \Return Anchors in $M$ sorted by their temporary coordinates
\EndFunction
\end{algorithmic}
\end{algorithm}

\section{Coarse- and Fine-Ranking Prompts}
\label{app:ranking_prompts}

The following templates reproduce the instructions used for coarse ranking
and local fine ranking, with runtime values replaced by placeholders.
In both prompts, \texttt{WEIGHTED\_RUBRICS} lists rubrics in the format
\texttt{1.[weight=5] rubric text}.
Anchors are indexed from 0 in the prompts: \texttt{A0} corresponds to
$a_1^t$ in the main text, and \texttt{A(K-1)} corresponds to $a_{K_t}^t$.

\subsection{Coarse-Ranking Prompt}

Here, \texttt{K} is the current number of anchors and \texttt{ORDERED\_ANCHOR\_BLOCKS} contains each anchor ID followed by its
response text, ordered from worst to best. The judge returns an insertion
position rather than a pointwise reward.

\begin{verbatim}
You are a strict response-quality judge. The anchors below are
already ordered from WORST to BEST.
Place the candidate response into exactly one insertion position
using the query and weighted rubrics.

Position semantics for {K} anchors:
- 0: candidate is worse than A0
- i (1 <= i < {K}): candidate belongs between A(i-1) and Ai
- {K}: candidate is better than A{K-1}

QUERY
{QUERY}

RUBRICS
{WEIGHTED_RUBRICS}

ORDERED ANCHORS (worst to best)
{ORDERED_ANCHOR_BLOCKS}

CANDIDATE
{CANDIDATE_RESPONSE}

Return JSON only: {"position": <integer from 0 through {K}>}
\end{verbatim}

\subsection{Fine-Ranking Prompt}
\label{app:listwise_prompt}

During fine ranking, this prompt is used for intervals containing at least
two rollouts.
\texttt{CANDIDATE\_BLOCKS} contains all rollouts in that interval, each
preceded by \texttt{[R<ID>]}. \texttt{CANDIDATE\_IDS} is the list of their
integer IDs. The returned \texttt{order} must contain each ID exactly once;
anchors are not included in this judgment.
Bootstrap initialization uses the same listwise prompt, with the initial
rollout group as candidates and no coarse ranking.

\begin{verbatim}
You are a strict response-quality judge. Rank only the candidate
responses below from WORST to BEST.
Use the query and weighted rubrics. Every candidate ID must
occur exactly once.

QUERY
{QUERY}

RUBRICS
{WEIGHTED_RUBRICS}

CANDIDATES
{CANDIDATE_BLOCKS}

Return JSON only. `order` must be a permutation of
{CANDIDATE_IDS}, from worst to best:
{"order": {CANDIDATE_IDS}}
\end{verbatim}

\section{Implementation Details}
\label{app:implementation_details}

Training uses one node with eight NVIDIA H800 GPUs, fully sharded data
parallelism for policy optimization, and vLLM for rollout generation.
Table~\ref{tab:implementation_hyperparameters} summarizes the training,
sampling, and buffer-maintenance settings. Unless otherwise specified,
these settings are shared across experiments; the initial anchor source,
buffer size, and maintenance protocol vary as described in the corresponding
experiments.

\begin{table}[htbp]
\caption{Main implementation settings.}
\label{tab:implementation_hyperparameters}
\begin{center}
\begin{tabular}{p{0.4\linewidth}p{0.48\linewidth}}
\toprule
Setting & Value \\
\midrule
\multicolumn{2}{l}{\textbf{Training}} \\
Queries per training step & 20 \\
Rollouts per query ($G$) & 8 \\
Policy learning rate & $1\times10^{-6}$ \\
Maximum prompt length & 4,096 tokens \\
Maximum response length & 4,096 tokens \\
Format-reward coefficient ($\lambda_{\mathrm{fmt}}$) & 0.2 \\
\midrule
\multicolumn{2}{l}{\textbf{Sampling and judging}} \\
Policy temperature & 0.6 \\
Policy top-$p$ & 0.95 \\
Policy top-$k$ & 20 \\
Judge decoding & Deterministic \\
\midrule
\multicolumn{2}{l}{\textbf{Rule-based buffer maintenance}} \\
Default initial buffer size ($K_0$) & 4 \\
Interval-refinement threshold & $|\mathcal{C}_p^t|\geq 3$ \\
Inactivity-pruning threshold ($H$) & 3 epochs \\

\bottomrule
\end{tabular}
\end{center}
\end{table}
\section{Rubric Generation}
\label{app:rubric_generation}

We generate query-specific rubrics using
DeepSeek-V4-Flash~\citep{xu2026deepseek} before RL training.
For each query, the model receives the query and its seven Nectar responses
in best-to-worst order and generates three to five concise rubrics.
Each rubric is assigned an integer importance weight from 1 to 5.
We use a temperature of 0.2 and a maximum output length of 1,024 tokens.
The generated output is parsed as JSON and validated to ensure that the number of rubrics and their weights satisfy the required constraints.
All successfully generated rubrics are kept fixed throughout RL training.

The complete system and user prompts are provided below.

\subsection{System Prompt}

\begin{verbatim}
You are an expert evaluator for assistant responses.
Given one user query and seven responses that are already sorted
from best to worst, write 3 to 5 concise judging rubrics that
explain the key dimensions that separate better answers from
worse ones for this specific query.

Requirements:
- The rubrics must be specific to this query and these seven
  responses.
- The rubrics should be useful for a later judge model to
  compare new responses.
- Focus on high-level evaluation dimensions such as correctness,
  completeness, instruction following, clarity, reasoning
  quality, safety, and style when relevant.
- Avoid mentioning the original seven answers explicitly in the
  rubric wording.
- Avoid redundant rubrics.
- Each rubric should be one sentence or short phrase.
- Assign each rubric an integer weight from 1 to 5.
- Larger weights mean the rubric is more important for judging
  future answers to this query.
- Use weight 5 only for the most critical dimensions.
- Output valid JSON only.

Return JSON in the following format:
{
  "rubrics": [
    {
      "text": "rubric 1",
      "weight": 5
    },
    {
      "text": "rubric 2",
      "weight": 4
    }
  ]
}
\end{verbatim}

\subsection{User Prompt}

\begin{verbatim}
Below is one query and seven candidate answers that are already
ranked from best to worst.

Your task:
1. Infer what makes the higher-ranked answers better than the
   lower-ranked ones.
2. Write 3 to 5 rubrics that can later be used to judge newly
   generated answers for the same query.
3. Assign each rubric an importance weight from 1 to 5, where 5
   means most important.
4. The rubrics should be general enough to evaluate future
   answers, but grounded in the distinctions visible in these
   seven ranked answers.

Query:
{QUERY}

Seven ranked answers (best to worst):

[Rank 1 | model={MODEL_1}]
{ANSWER_1}

[Rank 2 | model={MODEL_2}]
{ANSWER_2}

[Rank 3 | model={MODEL_3}]
{ANSWER_3}

[Rank 4 | model={MODEL_4}]
{ANSWER_4}

[Rank 5 | model={MODEL_5}]
{ANSWER_5}

[Rank 6 | model={MODEL_6}]
{ANSWER_6}

[Rank 7 | model={MODEL_7}]
{ANSWER_7}
\end{verbatim}

\section{Baseline Implementation Details}
\label{app:baseline_details}

\paragraph{Pointwise reward methods.}
For GRPO, Dr.\ GRPO, DAPO and GPG, each valid rollout is independently
scored once by the judge on a $0$--$10$ scale using the query-specific
weighted rubrics. The score $s$ is mapped to $2s/10-1$ before adding the
format reward. GDPO separately evaluates each rubric and normalizes the
rubric-specific rewards before aggregation, resulting in more judge calls.

\paragraph{Tournament and RRC methods.}
Tournament-GRPO runs three independently shuffled binary tournaments for
each group of eight rollouts, requiring 21 logical comparisons when all
requests succeed. RRC-SCR compares every unordered pair
in the group. RRC-AGR compares each rollout against the seven fixed Nectar
responses for its query. This is a reproducible adaptation because the
RRC paper specifies anchors sampled from a reference policy but does not
fix a public reference checkpoint; Nectar ranks are not exposed to the
AGR reward.

\paragraph{Exhaustive pairwise ranking.}
\label{app:pairwise_rank}
Pairwise Rank judges every unordered pair of non-empty responses;
a non-empty response automatically wins against an empty one.
We rank non-empty responses by decreasing win rate over completed
comparisons. Equal win rates share the first rank occupied by the tied
responses (e.g., $1,1,3$).
For $n$ non-empty responses, rank $s_i$ (1 is best) is mapped to
$2(n-s_i)/(n-1)-1$ for $n>1$ and $0$ for $n=1$, before adding the format
reward. Empty responses receive a semantic reward of $-1$.

\paragraph{ArenaRL.}
Our ArenaRL implementation uses the first rollout as a deterministic
seed response, compares each sampled rollout against that seed, and then
runs a seeded single-elimination tournament. Each logical match is judged
in both candidate orders. A group of eight therefore requires 14 logical
matches and 28 successful judge requests under error-free execution.

Across the ranking-based reward baselines, candidate presentation order is randomized
with a deterministic hash of the global seed, query, training step, and
candidate identifiers. Malformed or failed judge responses are retried;
only successfully parsed requests contribute to the reported call and
token totals.

\section{Complete Numerical Comparison}
\label{app:full_results}

Table~\ref{tab:all_results} reports the values underlying
Figures~\ref{fig:reward_efficiency} and~\ref{fig:reward_efficiency_calls}.
Judge tokens count prompt and completion
tokens used for reward construction during training; calls count
successfully parsed judge requests. The four benchmark columns use the
controlled evaluation protocol described in Section~\ref{sec:experimental_setup}.

\begin{table}[htbp]
\caption{Complete effectiveness and judge-cost comparison. An asterisk (*) denotes static buffers.}
\label{tab:all_results}
\begin{center}
\begin{tabular}{lrrrrrrr}
\toprule
Method & AE2 & AH-v2 & WB-v2 & Writing & Avg. & \shortstack{Tokens} & \shortstack{Calls} \\
\midrule
GRPO             & 72.73 & 27.50 & 46.57 & 65.42 & 53.06 & 25,035,966 & 31,374 \\
GDPO             & 64.35 & 20.10 & 45.14 & 64.94 & 48.63 & 86,937,684 & 144,236 \\
Dr.\ GRPO       & 56.15 & 18.90 & 44.32 & 63.37 & 45.60 & 21,936,802 & 31,649 \\
DAPO             & 74.22 & 26.60 & 46.83 & 66.13 & 53.44 & 24,078,025 & 31,231 \\
GPG              & 53.04 & 16.90 & 45.78 & 64.98 & 45.18 & 20,915,602 & 31,619 \\
\midrule
Tournament-GRPO  & 80.75 & 29.50 & 48.87 & 68.16 & 56.82 & 157,313,284 & 84,000 \\
RRC-AGR          & 71.74 & 23.30 & 46.61 & 64.79 & 51.61 & 200,817,235 & 218,914 \\
RRC-SCR          & 82.92 & 29.80 & 49.41 & 68.70 & 57.71 & 200,817,235 & 104,028 \\
Pairwise Rank    & 81.80 & 27.00 & 47.88 & 68.83 & 56.38 & 188,264,527 & 103,035 \\
ArenaRL          & 82.24 & 30.30 & 49.87 & 67.48 & 57.47 & 154,935,508 & 101,388 \\
\midrule
RankBuffer & 82.80 & 30.20 & 48.79 & 67.27 & 57.27 & 88,818,250 & 36,861 \\
RankBuffer* & 79.69 & 26.70 & 46.95 & 66.71 & 55.01 & 67,967,890 & 36,227 \\
\shortstack[l]{RankBuffer-\\Bootstrap} & 79.50 & 28.40 & 47.31 & 66.15 & 55.34 & 76,454,821 & 28,378 \\
\shortstack[l]{RankBuffer-\\Bootstrap*} & 81.61 & 29.60 & 47.47 & 67.48 & 56.54 & 60,662,185 & 28,108 \\
LLM & 81.74 & 30.50 & 47.34 & 66.83 & 56.60 & 96,503,178 & 41,592 \\
\shortstack[l]{LLM +\\look-ahead} & 79.07 & 28.40 & 46.31 & 66.13 & 54.98 & 110,997,178 & 42,101 \\
\bottomrule
\end{tabular}
\end{center}
\end{table}

\section{LLM Buffer Variants}
\label{app:llm_buffer}

The LLM variant starts from the same four Nectar
anchors as RankBuffer. After the current group has been ranked, the
manager receives the query, weighted rubrics, four current anchors,
current rollouts, and their combined worst-to-best ordering. It must
select exactly four responses from the union of old anchors and current
rollouts. The selected responses are copied verbatim and sorted according
to the established order to form the next buffer. Invalid outputs leave
the buffer unchanged.

The LLM + look-ahead variant delays this update until the query is encountered
in the next epoch. The manager additionally sees that epoch's newly
sampled, unranked rollouts as context, but may select only from the old
anchors and the previous epoch's ranked candidates. After selection, the
resulting buffer is used to rank the new rollouts.

The core selection instruction is:

\begin{verbatim}
Select the members of the next fixed-size reference buffer.
The buffer is a quality scale, not a collection of only the best
answers. Keep a stable, non-redundant set that covers the
response quality spectrum and meaningful transitions under the
rubrics.
Prefer a rollout when it fills a gap or is a better
representative than a current anchor.

Return valid JSON with exactly four selected IDs:
{
  "keep_anchor_ids": [IDs of retained current anchors],
  "add_rollout_ids": [IDs of selected eligible rollouts]
}
The two lists must contain four IDs in total. IDs not returned
are removed or discarded. Do not select look-ahead rollouts.
\end{verbatim}

\section{Ablation Configuration Details}
\label{app:ablation_details}

\paragraph{Initial-anchor subsets.}
For Nectar initialization, the seven responses are first ordered from worst
to best and then uniformly subsampled while preserving both endpoints. The
selected Nectar ranks are $(7,1)$ for $K_0=2$, $(7,5,3,1)$ for $K_0=4$, and
$(7,6,5,4,3,2,1)$ for $K_0=7$. For bootstrap initialization, we first rank the
eight rollouts sampled for each query in the first epoch from worst to best and
apply the same endpoint-preserving uniform subsampling. The selected positions
in that ordered group are $(1,8)$ for $K_0=2$, $(1,3,6,8)$ for $K_0=4$, and
$(1,2,3,5,6,7,8)$ for $K_0=7$. Subsequent buffer expansion, refinement, and
pruning are unchanged across the six variants.

\paragraph{Order-only configuration and judge prompt.}
The order-only control keeps the query, weighted rubrics, candidate response,
and the existing worst-to-best sequence of anchor identifiers, but removes the
anchor response text. Local fine ranking is otherwise unchanged. Its
coarse-ranking prompt is:

\begin{verbatim}
You are a strict response-quality judge. The anchors below are
already ordered from WORST to BEST. Place the candidate response
into exactly one insertion position using the query and weighted
rubrics.

Position semantics for K anchors:
- 0: candidate is worse than A0
- i (1 <= i < K): candidate belongs between A(i-1) and Ai
- K: candidate is better than A(K-1)

QUERY
{QUERY}

RUBRICS
{WEIGHTED_RUBRICS}

ORDERED ANCHORS (worst to best)
[A0]
...
[A(K-1)]

CANDIDATE
{CANDIDATE_RESPONSE}

Return JSON only: {"position": <integer from 0 through K>}
\end{verbatim}

\end{document}